# Evaluating structure learning algorithms with a balanced scoring function

Anthony C. Constantinou[1,2]

1. Bayesian Artificial Intelligence research lab, Risk and Information Management (RIM) research group, School of EECS, Queen Mary University of London (QMUL), London, UK, E1 4NS.
   E-mail: a.constantinou@qmul.ac.uk

2. The Alan Turing Institute, British Library, 96 Euston Road, London, UK, NW1 2DB.

**ABSTRACT:** Several structure learning algorithms have been proposed towards discovering causal or Bayesian Network (BN) graphs. The validity of these algorithms tends to be evaluated by assessing the relationship between the learnt and the ground truth graph. However, there is no agreed scoring metric to determine this relationship. Moreover, this paper shows that some of the commonly used metrics tend to be biased in favour of graphs that minimise edges. While graphs that are less complex are desirable, some of the metrics favour underfitted graphs, thereby encouraging limited propagation of evidence. This paper proposes the Balanced Scoring Function (BSF) that eliminates this bias by adjusting the reward function based on the difficulty of discovering an edge, or no edge, proportional to their occurrence rate in the ground truth graph. The BSF score can be used in conjunction with other traditional metrics to provide an alternative and unbiased assessment about the capability of a structure learning algorithm in discovering causal or BN graphs.

## 1. Introduction

Causality is the process by which an event, the cause, connects to another event, the effect, where the former is understood to be at least partly responsible for the latter. A Bayesian Network (BN) can model relationships under the assumption the arcs between nodes represent direct causation, often referred to as a Causal BN (CBN). The causal assumption is useful in cases where we seek to simulate the effect of intervention or to perform counterfactual reasoning (Pearl, 2013). Otherwise, a BN is viewed as a conditional dependency model amongst variables of interest.

Constructing a BN model involves determining its graphical structure which specifies the dependencies between variables, and its Conditional Probability Distributions (CPDs) which specify the relationship between directly dependent variables. Over the past three decades, several algorithms have been proposed that learn the structure of a BN from data. These algorithms largely fall into two categories, known as score-based and constraint-based learning.

Score-based learning represents a traditional machine learning approach that entails a search method for exploring the search space of graphs and an objective function for assessing each graph being explored. The graph that maximises the objective function is the preferred graph. Well-established algorithms that fall within this class of learning include the K2 (Cooper and Herskovits, 1992), the GES (Chickering, 2002), the Sparse Candidate (Friedman, et al., 1999), the ILP (Cussens, 2011), and the Optimal Reinsertion (Moore and Wong, 2003).

On the other hand, constraint-based learning involves assessing local sets of three nodes (or more) by performing conditional independence tests. The results from these tests determine the edges and the orientation of some of those edges. The PC (Spirtes, et al., 2000) and IC (Verma and Pearl, 1990) algorithms were the first to introduce this learning strategy, and the PC algorithm has become particularly popular in this area of research. Hybrid algorithms that combine these two classes of learning also exist, and include the L1-Regularisation paths (Schmidt, et al., 2007) and the Max-Min Hill-Climbing (MMHC; Tsamardinos, et al., 2006).

Determining the structure of a BN is not exclusively a machine learning problem. While structure learning is useful in disciplines such as bioinformatics where automated discovery may reveal insights that would otherwise remain unknown (Sachs, et al., 2005; Needham, et al., 2007; Maathuis, et al., 2009; Lagani, et al., 2016), knowledge-based graphs represent a common alternative approach in areas with access to expertise or rule-based information not captured by data. Some examples include medical applications (Coid, et al., 2016), law and forensics (Fenton, et al., 2016; Smit, et al., 2016), property market (Constantinou and Fenton, 2017), project management (Yet, et al., 2016), amongst many others (Fenton and Neil, 2018). As a result, structure learning algorithms often enable end users to incorporate knowledge-based constraints that determine what can and cannot be discovered by the algorithm; effectively representing a combination of the two approaches.

This paper focuses on the problem of evaluating the accuracy of a learnt graph by determining its relationship to the ground truth graph, independent of the type of parameter learning that could be used to parameterise the CPDs of the graph. Hence, the focus of this paper is fully oriented towards graphical discovery rather than inference. The remainder of the paper is structured as follows: Section 2 reviews the commonly used approaches to evaluate learnt structures, Section 3 justifies the need for a balanced score, Section 4 describes the novel Balanced Scoring Function (BSF), Section 5 provides hypothetical and experimental analysis to highlight the difference between the proposed metric and the traditional metrics, and Section 6 provides the concluding remarks.

## 2. Scoring metrics

Various metrics have been considered in the literature to evaluate the accuracy of a learnt graph with respect to the ground truth graph. While the ground truth graph is generally assumed to be a Directed Acyclic Graph (DAG), it is occasionally assumed to be a Complete Partial Directed Acyclic Graph (CPDAG[1]) which incorporates both directed and undirected edges. A CPDAG represents a Markov Equivalence class of DAGs, which means that data simulated by the DAGs that falls within that class is indistinguishable. In this paper, however, we shall focus on the assessment of DAGs since they entail more information than CPDAGs and tend to be preferred when applying BNs to real-world problems. Regardless, any conclusions about the metrics derived from the assessment of DAGs are expected to apply to CPDAGs.

[1] Also called PDAGs by Spirtes and Meek (1995), *patterns* by Spirtes et al (Spirtes, et al., 2000), *essential graphs* by Andersson et al (Andersson, et al., 1997), and *maximally oriented graphs* by Meek (Meek, 1995).

### 2.1. Metrics based on derivations from a confusion matrix

For the purposes of BN structure learning, a confusion matrix is a simple two-dimensional table that captures a set of parameters that tell us something about the relationship of a learnt graph with respect to the ground truth graph. Table 1 presents a hypothetical example based on the simplest possible confusion matrix that could be considered where:

i. *True Positives* (TP) corresponds to the number of true arcs present in the learnt graph,
ii. *False Positives* (FP) corresponds to the number of false arcs present in the learnt graph,
iii. *True Negatives* (TN): corresponds to the number of true direct independencies present in the learnt graph,
iv. *False negatives* (FN): corresponds to the number of false direct independencies present in the learnt graph; i.e., arcs not discovered.

**Table 1.** A simple confusion matrix that could be considered by a scoring metric when evaluating the relationship between a learnt graph and the ground truth graph. The hypothetical example assumes a ground truth DAG with 10 nodes, 10 arcs, and a learnt DAG with 12 arcs.

| | **Ture arcs** | **True independencies** |
|---|---|---|
| Learnt arcs | 7 [TP] | 5 [FP] |
| Learnt independencies | 3 [FN] | 30 [TN] |

The example in Table 1 assumes a problem over a set of nodes $N$, and a graph with a variable set of size $|N| = 10$ that can have $j$ possible edges between each pair of nodes $n_i, n_j \in N$ where

$$j = \frac{|N|(|N| - 1)}{2} \quad (1)$$

Therefore, the confusion matrix parameters must satisfy $j = TP + FP + FN + TN$. In the example presented in Table 1, the confusion matrix parameters sum up to 45, where 45 corresponds to the maximum number of edges that could be produced for a network that incorporates 10 nodes, as described by equation 1.

Various derivations can be retrieved from a confusion matrix, and these can be confusing in their interpretation. The derivations discovered in the papers reviewed, listed in Table 2, are:

i. *Partial Confusion Matrix (PCM) stats*: the cases where the authors would report some of the parameters of the confusion matrix, and typically focus on TP, FP, or FN, or any combination of the three.

ii. *Precision*: the cases where the authors would report the Precision ($Pr$) rate, defined as

$$Pr = \frac{TP}{TP + FP} \quad (2)$$

and represents the rate of correct direct dependencies from those discovered. For example, the Precision rate in Table 1 is 58.3% because the hypothetical learnt graph discovered seven true arcs out of the 12 arcs learnt.

iii. *Recall (Re)*, also referred to as *Sensitivity*, represents the proportion of direct dependencies discovered from those in the true graph defined as

$$Re = \frac{TP}{TP + FN} \tag{3}$$

For example, the $Re$ rate in Table 1 is 70% because the hypothetical learnt graph discovered seven out of the 10 true arcs.

*2.2. Metrics based on structural discrepancies*

The metrics that measure structural discrepancies between graphs generally produce a score that corresponds to some difference between the learnt graph and the ground truth graph. One of the most commonly used metrics in this field of research is the Structural Hamming Distance (SHD) proposed by Tsamardinos et al (2006). The SHD score represents the minimum number of edge insertions, deletions, and arc reversals needed to convert the learnt graph into the true graph. It turns out that

$$SHD = FN + FP \tag{3}$$

where arc reversals fall either under $FN$ or $FP$.

Similarly, in (Constantinou, et al., 2019) the *DAG Dissimilarity Metric* (DDM) was used to determine the level of dissimilarity between two DAGs. The score ranges from $-\infty$ to 1, where a score of 1 indicates a perfect agreement between the two graphs. The score moves to $-\infty$ the stronger the dissimilarity is between the two graphs. Specifically, if we compare how dissimilar the learnt graph is with respect to true graph, then

$$DDM = \frac{TP + \frac{r}{2} - FN - FP}{a} \tag{4}$$

where $r$ is the number of arcs in the ground truth graph reoriented in the learnt graph, and $a$ is number of arcs in the true graph. Note that $r$ is equivalent to the number of arc reversals in SHD, although its penalty is halved in DDM to acknowledge the fact that an arc reversal is an event preceded by the successfully discovery of an edge. In other words, DDM acknowledges that an arc reversal corresponds to the discovery of a correct direct dependency without the correct orientation and which can be viewed as a partially correct (or a partially incorrect) arc.

It is also worth mentioning Peters and Buhlmann's (2014) Structural Intervention Distance (SID) metric which was proposed as an extension of SHD with a focus on measuring the causal effects in CBNs via intervention. Although a relatively recent metric, the restriction of SID to causal graphs may explain why SID has not been adopted by the relevant literature. Since this paper focuses on BN graphs that are not necessarily causal graphs, the SID metric is out of the scope of this paper.

### 2.3. *Inference methods used as scoring metrics*

The most common inference-based methods for BN model selection include the Bayesian Information Criterion (BIC), also known as the Maximum Description Length (MDL), and the Bayesian scores BD/BDe/BDeu with or without equivalent priors. While these scores often serve as the objective function in score-based learning, they also regularly serve as scoring metrics in determining the accuracy of a learnt graph.

Inference-based methods tell us how well a model may perform in terms of prediction. Because they are score-equivalent, however, they cannot distinguish between DAGs that fall under the same Markov equivalence class. Moreover, in their mission to balance model fitting with respect to model dimensionality, they remain oblivious to the dimensionality of the true graph. As a result, a higher model selection score does not always imply a more accurate graph (Constantinou et al., 2020). Likewise, inference-based scores are inherently biased in favour of the algorithms that maximise them. For example, if the evaluation metric is the BIC score, a score-based algorithm whose objective function maximises the BIC score will likely outperform other algorithms that optimise for some other objective function, including constraint-based algorithms. Because of these reasons, this paper focuses on metrics that are fully orientated towards graphical discovery rather than inference.

### 2.4. *Scoring metrics used in the literature*

The different types of evaluation of a leant graph with respect to the ground truth, or some gold standard, graph can be categorised as follows:

a) Type CM (Confusion Matrix) which involves scoring metrics that are based on derivations from a confusion matrix, as defined in subsection 2.1;
b) Type SD (Structural Discrepancy) which involves scoring metrics based on the number of structural discrepancies between graphs, as defined in subsection 2.2;
c) Type I (Inference) which involves inference-based methods as discussed in subsection 2.3;
d) Type T (Theory) which involves supporting the validity of a structure learning algorithm with theoretical proofs, but only when used as an alternative, and not in addition, to empirical evaluation.

Table 2 lists 137 papers reviewed and indicates the approaches used to evaluate graphical structures in each paper. The evaluation of graphical structures reviewed would generally involve determining the effectiveness of a structure learning algorithm, and occasionally this would include evaluating the effectiveness of knowledge-based constraints. The most commonly used evaluators are listed under each type of evaluation, where PCM is Partial Confusion Matrix stats such as TP and FP, OSD is Other Structural Discrepancies such as edges removed and edges added, AUC is the Area Under the Curve of the Receiving Operating Characteristic (ROC), and OI is any Other Inference-based measures such as Log-Likelihood, AIC, KL divergence and predictive accuracy. Note that theoretical proof (Type T) is reported only for papers that did not involve empirical validation and based the validity of an algorithm solely on theory.

**Table 2.** The evaluation approaches used in the relevant 137 papers reviewed, where type CM represents metrics based on derivations from a confusion matrix, type SD represents metrics based on structural discrepancies between graphs, type I represents inference-based methods, and type T represents cases where theoretical proof was used as an alternative (and *not* in addition) to empirical validation.

| | | Type CM | | | Type SD | | Type I | | | | Type T |
|---|---|---|---|---|---|---|---|---|---|---|---|
| **Year** | **Author/s** | **PCM** | ***Pr*** | ***Re*** | **SHD** | **OSD** | **BIC** | **BD/e/u** | **AUC** | **OI** | |
| 1968 | Chow and Liu | | | | | | | | | × | |
| 1990 | Verma and Pearl | | | | | | | | | | × |
| 1990 | Srinivas, et al. | | | | | × | | | | | |
| 1991 | Spirtes and Glymour | | | | | × | | | | | |
| 1992 | Cooper and Herskovits | × | | | | | | | | × | |
| 1994 | Bouckaert | | | | | | × | × | | | |
| 1995 | Chicheking | | | | | | | | | | × |
| 1995 | Singh and Valtorta | | | | | × | | | | | |
| 1995 | Spirtes, et al. | | | | | | | | | | × |
| 1995 | Meek | | | | | | | | | | × |
| 1995 | Heckerman, et al. | | | | | × | | | | × | |
| 1996 | Richardson | | | | | | | | | | × |
| 1996 | Larranaga, et al. | | | | | × | | | | × | |
| 1996 | Madigan, et al. | | | | | | | | | × | |
| 1997 | Etxeberria, et al. | | | | | × | | | | × | |
| 1999 | Wong, et al. | | | | | | × | | | | |
| 1999 | Friedman, et al. | | | | | | | × | | | |
| 1999 | Myers, et al. | | | | | | | × | | × | |
| 2000 | Tian | | | | | × | | | | | |
| 2000 | Castelo and Siebes | | | | | | | | | × | |
| 2002 | Cheng, et al. | | | | | × | | | | | |
| 2002 | Chickering | | | | | × | | × | | | |
| 2003 | Dash and Druzdzel | | | | | | | | | × | |
| 2003 | Moore and Wong | | | | | | | × | | | |
| 2003 | Friedman and Koller | × | | | | | | | | × | |
| 2003 | Castelo and Kocka | | | | | × | | × | | | |
| 2003 | Tsamardinos | | | | | | | | × | | |
| 2003 | Acid and de Campos | | | | | × | × | × | | × | |
| 2003 | Margaritis | × | | | | × | × | | | × | |
| 2004 | Koivisto and Sood | | | | | × | | | | | |
| 2004 | Dash and Cooper | | | | | | | | × | | |
| 2005 | Teyssier and Koller | | | | | | | × | | × | |
| 2005 | Yaramakala and Margaritis | | | | | | | | | × | |
| 2006 | Shimizu, et al. | | | | | × | | | | | |
| 2006 | Ramsey, et al. | | | | | × | | | | | |
| 2006 | Zuk, et al. | | | | | | | | | × | |
| 2006 | Meganck, et al. | | | | | | | | | × | |
| 2006 | Tsamardinos, et al. | | | | × | | | × | | | |
| 2007 | Eaton and Murphy | | | | | × | | | × | × | |
| 2007 | Kalisch and Buhlmann | × | | | × | | | | | | |
| 2007 | Schmidt, et al. | | | | | × | × | | | × | |
| 2008 | Chen, et al. | | | | | × | | | | | |
| 2008 | Perrier, et al. | | | | | × | × | | | × | |
| 2008 | Elidan and Gould | | | | | | | | | × | |
| 2008 | Ellis and Wong | | | | | | | | × | | |
| 2008 | Pellet and Elisseeff | | | | | × | | | | | |
| 2008 | Grzegorczyk and Husmeier | | | | | | | | × | × | |
| 2008 | Cussens | | | | | | | × | | × | |
| 2008 | Wong and Guo | | | | | | | | | × | |
| 2008 | Zhang | | | | | | | | | | × |
| 2008 | Xie and Geng | | | | × | × | | | | | |
| 2009 | Bromberg and Margaritis | | | | | | | | | × | |
| 2009 | Maathuis, et al. | × | | | | | | | | × | |
| 2009 | Sierra, et al. | | | | | | | | | × | |
| 2009 | Danks, et al. | | | | | × | | | | | |
| 2009 | Daly and Shen | | | | × | | | × | | | |
| 2009 | Pinto, et al. | | | | × | | | × | | | |
| 2009 | Yehezkel and Lerner | | | | × | × | | × | | | |
| 2010 | Bouchaala, et al. | × | | | | | | | | | |
| 2010 | Pernkopf | | | | | | | | | × | |
| 2010 | Aliferis, et al. | | | | | × | | | × | | |
| 2010 | Buhlmann, et al. | × | | | | | | | | × | |
| 2010 | Maathuis, et al. | | | | | × | | | × | | |
| 2010 | Studeny, et al. | | | | | | | | | | × |
| 2010 | de Morais and Aussem | | × | × | | × | | | | | |
| 2010 | Noble | | | | | × | | | | | |
| 2010 | Jaakkola | | | | | | | × | | | |
| 2010 | Wand and Yang | | | | | | | | | × | |
| 2011 | Alameddine, et al. | | | | | × | × | | | × | |
| 2011 | Flores, et al. | | | | | × | | | | | |
| 2011 | Ueno | | | | | | × | × | | | |
| 2011 | de Campos | | | | | | × | × | | | |
| 2011 | Yang, et al. | | | | | × | | | | | |
| 2011 | Alonso-Barba and delaOssa | | | | × | | | × | | | |
| 2011 | Smith, et al. | × | | | | × | | | | | |
| 2012 | Gasse, et al. | × | × | × | × | × | × | × | | | |
| 2012 | Colombo, et al. | | | | | × | | | | | |
| 2012 | Scutari and Brogini | | | | × | | × | × | | × | |
| 2013 | Zhang, et al. | | | | | | | × | | | |
| 2013 | Masegosa and Moral | × | | | × | × | | | | × | |
| 2013 | Huang, et al. | × | | | | | | | | | |
| 2013 | Barlett and Cussens | | | | | | | × | | | |
| 2013 | Niinimaki and Koivisto | | | | | | | | | × | |
| 2013 | Cano, et al. | × | | | | | | × | | × | |
| 2013 | Ji, et al. | | | | | | | | | × | |
| 2013 | Borboudakis and Tsamardinos | | | | × | | | | | × | |
| 2013 | Alonso-Barba, et al. | | | | × | × | × | | | | |
| 2013 | Brenner amd Sontag | | | | × | | | | | × | |
| 2013 | Harris and Drton | | | | | | | | × | × | |
| 2014 | Studeny and Haws | | | | | | × | × | | | |
| 2014 | Parviainen, et al. | | | | | | | × | | | |
| 2014 | Colombo and Maathuis | × | | | × | | | | × | | |
| 2015 | Suzuki | | | | | | | | | | × |
| 2015 | Scanagatta, et al. | | | | | | × | | | | |
| 2015 | Malone, et al. | | | | | | | | | × | |
| 2015 | Ramsey | | × | × | | | | | | | |
| 2016 | He | | | | | | | | | × | |
| 2016 | Goudie and Mukherjee | × | | | × | × | | | × | × | |
| 2016 | Gheisari and Meybodi | | | | | × | × | | | × | |
| 2016 | Pensar, et al. | × | | | | | | | | × | |
| 2016 | Yang, et al. | | | | | × | | | | × | |
| 2016 | Oates, et al. | | | | × | | | | | | |
| 2016 | Ramsey | | × | × | | | | | | | |
| 2016 | Scutari | | | | | | × | × | | × | |
| 2016 | Ogarrio, et al. | | × | × | | | | | | | |
| 2016 | Triantafillou and Tsamardinos | | × | × | × | | | | | | |
| 2017 | Djordjilovic | | | | × | | | | | × | |
| 2017 | Lee and van Beek | | | | | | × | × | | | |
| 2017 | Gao, et al. | | | | | | | × | | | |
| 2017 | Ji, et al. | | | | | × | | | | × | |
| 2017 | Nie, et al. | | | | | | | × | | × | |
| 2017 | Parviainen and Kaski | | | | × | | | | | | |
| 2017 | Suzuki | | | | | | | | | × | |
| 2017 | Tabar | × | | | | | | | | | |
| 2017 | Ramsey, et al. | | × | × | | | × | | | | |
| 2018 | Talvitie, et al. | | | | × | × | | | | | |
| 2018 | Sanchez-Romero, et al. | | × | × | | | | | | | |
| 2018 | Gao and Wei | | | | | | | × | | | |
| 2018 | Tsirlis, et al. | | × | × | × | × | | | | × | |
| 2018 | Chen, et al. | | | | | | × | | | | |
| 2018 | Li and van Beek | × | | | | × | | × | | | |
| 2018 | Scanagatta, et al. | | | | | | × | × | | × | |
| 2018 | Behjati and Beigy | | | | | | | × | | | |
| 2018 | Córdoba, et al. | | | | × | × | | | | | |
| 2018 | Silander, et al. | | | | | | × | × | | × | |
| 2018 | Nandy, et al. | × | | | | | | | | | |
| 2018 | Jennings Corcoran | | | | | | | | | × | |
| 2018 | Scutari, et al. | | | | × | | | | | | |
| 2018 | de Campos, et al. | | | | | | × | | | | |
| 2019 | Zhao and Ho | | | | × | × | × | × | | | |
| 2019 | Eggeling, et al. | × | | | × | | × | × | | × | |
| 2019 | Liao, et al. | | | | | | × | × | | | |
| 2019 | Yu, et al. | × | | | × | | × | | | | |
| 2020 | Constantinou, et al. | | × | × | × | × | | | | | |
| 2020 | Kiattikun and Constantinou | | × | × | × | × | | | | | |
| 2020 | Correia, et al. | | | | | | | | | × | |
| 2020 | Kitson and Constantinou | × | × | × | × | × | × | | | × | |

Table 2 suggests that approaches based on structural discrepancies and inference have always been common in evaluating graphical structures, whereas metrics based on deviations from a confusion matrix have become common over last decade. On the other hand, theoretical proofs were used as an alternative to empirical validation mostly during the first few years, and have only been used occasionally since then (although they are still often used in addition to empirical validation). Specifically, out of the 137 papers reviewed, 21 papers reported results based on partial stats derived from a confusion matrix (e.g., TP, FP), 12 papers reported the Precision and Recall, 29 papers reported the SHD score, 47 papers reported a score based on some structural discrepancy score but which was not the SHD score (e.g., edges removed/added), 27 papers reported the BIC (including MDL) score, 36 papers reported any of the Bayesian scores BD/e/u, 10 papers reported the AUC of ROC, 56 papers reported any other type of inference (e.g., LL, KL-divergence, AIC, predictive validation), and eight papers were solely based on theoretical proof. Overall, 69 out of the 137 papers based the evaluation on a single evaluation approach from those covered in Table 2, 39 papers on two approaches, 15 papers on three approaches, and 10 papers on four or more approaches.

## 3. The need for a balanced score

Section 2 described how each of the metrics, that are not based on inference, relate to the parameters of the confusion matrix. Table 3 presents the scores generated by the different metrics based on the hypothetical confusion matrix parameters TP, FP, TN and FN. The hypothetical scenarios enumerated in Table 3 assume that the true graph is a DAG which consists of 10 variables and 10 arcs, and that each learnt graph is also a DAG with varying numbers of arcs.

**Table 3.** The scores produced by each metric given the hypothetical TP, FP, TN, and FN confusion matrix parameters. The illustration assumes that the true DAG consists of 10 variables and 10 arcs.

| Scenario | Scenario description | TP | FP | TN | FN | Pr | Re | F1 | SHD | DDM |
|---|---|---|---|---|---|---|---|---|---|---|
| 1.1 | Discrepancies in TP | 10 | 20 | 15 | 0 | 0.33 | 1 | 0.5 | 20 | -1 |
| 1.2 | (and consequently FN) | 5 | 20 | 15 | 5 | 0.2 | 0.5 | 0.29 | 25 | -2 |
| 1.3 | | 0 | 20 | 15 | 10 | 0 | 0 | n/a | 30 | -3 |
| 2.1 | Discrepancies in FP | 5 | 15 | 20 | 5 | 0.25 | 0.5 | 0.33 | 20 | -1.5 |
| 2.2 | (and consequently TN) | 5 | 10 | 25 | 5 | 0.33 | 0.5 | 0.4 | 15 | -1 |
| 2.3 | | 5 | 5 | 30 | 5 | 0.5 | 0.5 | 0.5 | 10 | -0.5 |
| 3.1 | Fully connected graph A | 10 | 35 | 0 | 0 | 0.22 | 1 | 0.36 | 35 | -2.5 |
| 3.2 | Fully connected graph B | 5 | 35 | 0 | 5 | 0.125 | 0.5 | 0.2 | 40 | -3.5 |
| 3.3 | Empty graph | 0 | 0 | 35 | 10 | n/a | 0 | n/a | 10 | -1 |
| 3.4 | Most inaccurate graph | 0 | 35 | 0 | 10 | 0 | 0 | n/a | 45 | -4.5 |
| 3.5 | Most accurate graph | 10 | 0 | 35 | 0 | 1 | 1 | 1 | 0 | 1 |

The illustration is based on three different sets of scenarios. In the first set, TP is the only parameter that varies, but also FN as a consequence of TP. In the second set, FP is the only parameter that varies, but also TN as a consequence of FP. Lastly, the third set consists of the following scenarios:

i. a fully connected graph ($A$) in which all correct edges have the correct orientation;
ii. a fully connected graph ($B$) in which half of the correct edges have the correct orientation, representing the case of a randomly generated fully connected graph;
iii. an empty graph with no edges;
iv. the most inaccurate graph given the true graph, where the learnt graph has an arc between nodes that have no arc in the true graph, and no arcs between nodes that have an arc in the true graph;
v. the most accurate graph, where the learnt graph is a perfect match of the true graph.

To begin with, it should be clear why the TP, FP, TN, and FN parameters must not be used independently to judge a learnt graph. As shown in Table 3, this is because TP remains invariant when FP varies, and FP and TN remain invariant when TP varies. As obvious as this may seem, statistics related to TP, FP and TN parameters are commonly used to judge a learnt graph. Although these statistical indicators are often reported in addition to some other metric or evaluation approach, the level of emphasis put on some of these indicators varies across papers. While it may be reasonable to refer to these statistical indicators independently by way of *explanation* as to why two graphs may differ, they should not independently form part of the assessment process.

Furthermore, while Precision and Recall are commonly used throughout the scientific literature, their use has been relatively limited in this area of research. This is because they tend to be more popular with constraint-based, rather than score-based, algorithms. Regardless, these metrics are also inadequate when used independently and this is because each of these metrics will only consider two out of the four confusion matrix parameters. Their limitations are well documented in other research fields, and this has led to the introduction of the F1 score which is viewed as a 'fairer' metric that provides the harmonic mean of Precision and Recall:

$$\mathrm{F1} = 2\frac{Pr.Re}{Pr + Re} \qquad (5)$$

Only a handful of the papers reviewed have considered the F1 score in their assessment (recorded under PCS). Still, the F1 score also has its limitations, which are also well documented in other disciplines (Hand and Christen, 2018). For the purposes of this paper, the limitations of the F1 score include a) it does not take into account the TN parameter, b) it assumes that Precision and Recall are equally important, and c) the F1 score cannot be computed when the learnt graph does not capture one of the true arcs (refer to Table 3), which can often occur when working with networks that incorporate a low number of nodes.

Lastly, the SHD and DDM metrics address some of the limitations discussed above and are able to correctly order each of the learnt graphs within each of the three scenarios in Table 3. However, note how both metrics strongly favour the empty graph over most of the other graphs. Yet, an empty graph is useless since it provides zero propagation of evidence. Importantly, the score assigned to the empty graph not only is considerably higher than the score assigned to the fully connected graphs, which can be viewed as an equally poor attempt in reconstructing the true graph, but it is also the shared best score over all graphs and over all scenarios excluding the scores assigned to the 'Most accurate graph'.

This observation exposes a clear bias for measures that are based on structural discrepancies, including those reported under OSD. It turns out the SHD score, and in general any other structural discrepancy-based metric, represents a score that merely approximates classic classification accuracy. For example, if the true graph has 1% arcs and 99% direct independencies, a structural discrepancy-based metric would judge an empty graph as being up to 99% accurate in relation to the true graph on the basis that all of the direct independencies have been discovered, which represent up to 99% of the network. While classification accuracy is widely considered to be misleading in other fields of machine learning research, the simplicity and transparency of structural discrepancy-based measures are perhaps the reasons why they remain popular in this field of research.

## 4. The Balanced Scoring Function (BSF)

Since a scoring metric is based on the parameters of a confusion matrix (or similar), then the definition of those parameters also influences the outcome of the metric. For example, the confusion matrix described in Section 2.1 does not differentiate between the errors of a) incorrect dependency and b) correct dependency with an incorrect orientation; i.e., these errors would generally fall under FP or FN. This type of error extends to the SHD metric and any other metric that assumes that the penalty of discovering a false dependency is equivalent to the penalty of discovering a true dependency but with the incorrect direction (e.g., the SHD error would be 1 for both cases). Tsamardinos et al (2006) acknowledged that the SHD is biased towards the sensitivity of identifying edges versus specificity. Colombo and Maathuis (2014) tried to improve the SHD by introducing a variant called the *SHD edge marks* that explicitly counts the number of errors in edge marks by penalising the change from A ↔ B to A → B by 1, and the change from A → B to A ⊥ B by 2. Similarly, DDM metric assumes that the reward of discovering a true dependency without the correct orientation is half relative to the reward of discovering a dependency with the correct orientation. For a detailed comparison of the different properties of the structural discrepancy metrics see (de Jongh and Druzdzel, 2009).

### *4.1. Determining the penalty function*

Table 4 presents the proposed penalty weights that can serve as the basis for any scoring metric. These weights assume that if $A \rightarrow B$ exists in the true graph, then the undirected edge $A - B$, the bi-directed edge A ↔ B, and the reversed arc A ← B would be considered as a 'Partial match' in a learnt graph. In other words, the proposed penalty weights assume that a 'Partial match' corresponds to the discovery of the true dependency without discovering the correct direction of that dependency. A possible argument here is that the penalty for reversing an arc should be higher than the penalty of directing an undirected or a bi-directed edge, similar to what Colombo and Maathuis (2014) proposed. However, a counterargument is that we should not be rewarding algorithms that make no attempt to determine the orientation of an edge higher than we reward those that proceed to orientate them.

**Table 4.** Proposed penalty weights for the any scoring metric. Rule 6 applies only to Markov equivalent DAGs (i.e., CPDAGs).

| Rule | True graph | Learnt graph | Penalty | Justification |
|---|---|---|---|---|
| 1 | A → B | A → B | 0 | Complete match |
| 2 | A → B | A ↔ B, A − B or A ← B | 0.5 | Partial match |
| 3 | A → B | A ⊥ B | 1 | No match |
| 4 | A ⊥ B | A ⊥ B | 0 | Complete match |
| 5 | A ⊥ B | Any edge/arc | 1 | Incorrect dependency discovered |
| 6 | A → B | Any edge/arc | 0 | Only if Markov equivalent |

### *4.2. Eliminating the bias between direct dependencies and direct independencies*

Under the assumption that the difficulty of discovering an arc is directly dependent on the number of arcs that exist in the true graph, relative to the number of direct independencies, a balanced scoring metric would need to weight the difficulty of discovering arcs $w_a$ and independencies $w_i$ as follows:

$$w_a = \frac{1}{a}, \ w_i = \frac{1}{i} \qquad (6)$$

where $a$ and $i$ represent the number of arcs and independencies in the true graph respectively. If the number of arcs in the true graph is $a$, then

$$i = \frac{|N| \times (|N| - 1)}{2} - a \quad (7)$$

or simply $i = j - a$ (given equation 1), where $|N|$ is the size of node set $N$ as defined in subsection 2.1. For example, in Table 3, $a = 10$ and $i = 45 - 10 = 35$. In that example, $a$ is 3.5 times more prevalent that $i$ and hence, $w_a = \frac{1}{10}$ relative to $w_i = \frac{1}{35}$. A balanced score $BS$ requires that we adjust each parameter $k$ in the confusion matrix to its prevalence rate as follows:

$$BS = \frac{2}{k} \sum_k \begin{cases} k \times w_a, & k = TP \\ k \times w_i, & k = TN \\ -k \times w_i, & k = FP \\ -k \times w_a, & k = FN \end{cases} \quad (8)$$

This also ensures the score will range from -1 to 1, where a score of -1 corresponds to the least accurate graph, a score of 1 to a graph that is a perfect match of the true graph, and a score of 0 to a graph that is no better or worse than a baseline empty or a fully connected graph. Thus, the score of the Balanced Scoring Function (BSF) can be obtained using equation 9:

$$\mathrm{BSF} = \left(\frac{\mathrm{TP}}{a} + \frac{\mathrm{TN}}{i} - \frac{\mathrm{FP}}{i} - \frac{\mathrm{FN}}{a}\right) \Big/ 2 \quad (9)$$

## 5. Hypothetical and empirical analysis

### *5.1. Hypothetical analysis*

Table 5 represents a revised version of Table 3 where the metrics are adjusted to consider the 'Partial match' (indicated as $TP^b$) rule as part of the universal penalty weights proposed in Table 4, and to include the scores generated by the novel BSF. The addition of 'partial arcs discovered', which now carry half[2] the weight of TP, have not led to any significant changes compared to the results in Table 3. Moreover, the BSF scores are in agreement with the SHD and DDM metrics in terms of ranking the learnt graphs by score within each set of scenarios. However, there are clear disagreements when comparing the metric scores across scenarios from different sets.

[2] For example, the total TP score for scenario is $TP + {TP^b}/{2} = 9$.

**Table 5.** The scores from Table 3 revised to account for the 'Partial match' as defined in Table 4, and the scores produced by the BSF.

| Scenario | Scenario description | TP | TP^b | FP | TN | FN | Pr | Re | F1 | SHD | DDM | BSF |
|---|---|---|---|---|---|---|---|---|---|---|---|---|
| 1.1 | Discrepancies in TP | 8 | 2 | 20 | 15 | 1 | 0.3 | 0.9 | 0.45 | 21 | -1.2 | 0.329 |
| 1.2 | (and consequently FN) | 4 | 1 | 20 | 15 | 5.5 | 0.18 | 0.45 | 0.257 | 25.5 | -2.1 | -0.121 |
| 1.3 | | 0 | 0 | 20 | 15 | 10 | 0 | 0 | n/a | 30 | -3 | -0.571 |
| 2.1 | Discrepancies in FP | 4 | 1 | 15 | 20 | 5.5 | 0.225 | 0.45 | 0.3 | 20.5 | -1.6 | 0.021 |
| 2.2 | (and consequently TN) | 4 | 1 | 10 | 25 | 5.5 | 0.3 | 0.45 | 0.36 | 15.5 | -1.1 | 0.164 |
| 2.3 | | 4 | 1 | 5 | 30 | 5.5 | 0.45 | 0.45 | 0.45 | 10.5 | -0.6 | 0.307 |
| 3.1 | Fully connected graph A | 10 | 0 | 35 | 0 | 0 | 0.22 | 1 | 0.364 | 35 | -2.5 | 0 |
| 3.2 | Fully connected graph B | 5 | 5 | 35 | 0 | 2.5 | 0.167 | 0.75 | 0.273 | 37.5 | -3 | -0.25 |
| 3.3 | Empty graph | 0 | 0 | 0 | 35 | 10 | n/a | 0 | n/a | 10 | -1 | 0 |
| 3.4 | Most inaccurate graph | 0 | 0 | 35 | 0 | 10 | 0 | 0 | n/a | 45 | -4.5 | -1 |
| 3.5 | Most accurate graph | 10 | 0 | 0 | 35 | 0 | 1 | 1 | 1 | 0 | 1 | 1 |

Fig 1 illustrates the disagreements between BSF and the other metrics by plotting the BSF scores generated for each of the scenarios in Table 5, and ordering the scenarios by BSF score before superimposing the scores generated by the other metrics. Therefore, the scenarios are ordered from the most to least accurate, as determined by the BSF. Since the scores produced by the different metrics differ in score range as well as in terms of what a higher score represents, the scores had to be normalised before comparing them.

Fig 1 illustrates how the scores compare to the BSF scores after normalising them between 0 and 1 and adjusting them so that 1 represents the highest accuracy and 0 the lowest accuracy across all metrics. The $Pr$, $Re$, and F1 scores already satisfy the above definition and required no normalisation. However, cases where the F1 score produced 'n/a' assume a score of 0. The rest of the metric scores are normalised as follows:

$$BSF_N = \frac{BSF + 1}{2} \tag{10}$$

where $BSF_N$ is the BSF score normalised between 0 and 1,

$$SHD_N = 1 - \frac{SHD}{\max_{G_i}(SHD)} \tag{11}$$

where $\max_{G_i}(SHD)$ is the highest SHD score generated over all the learnt graphs that associate with the ground truth graph $G_i$, and $SHD_N$ is the SHD score normalised between 0 and 1 and reversed so that a higher SHD score indicates a better performance, and

$$DDM_N = \frac{DDM + \left|\min_{G_i}(DDM)\right|}{\left|\min_{G_i}(DDM)\right| + 1} \tag{12}$$

where $\min_{G_i}(DDM)$ is the lowest DDM score generated over all the learnt graphs that associate with the ground truth graph $G_i$, and $DDM_N$ is the DDM score normalised between 0 and 1.

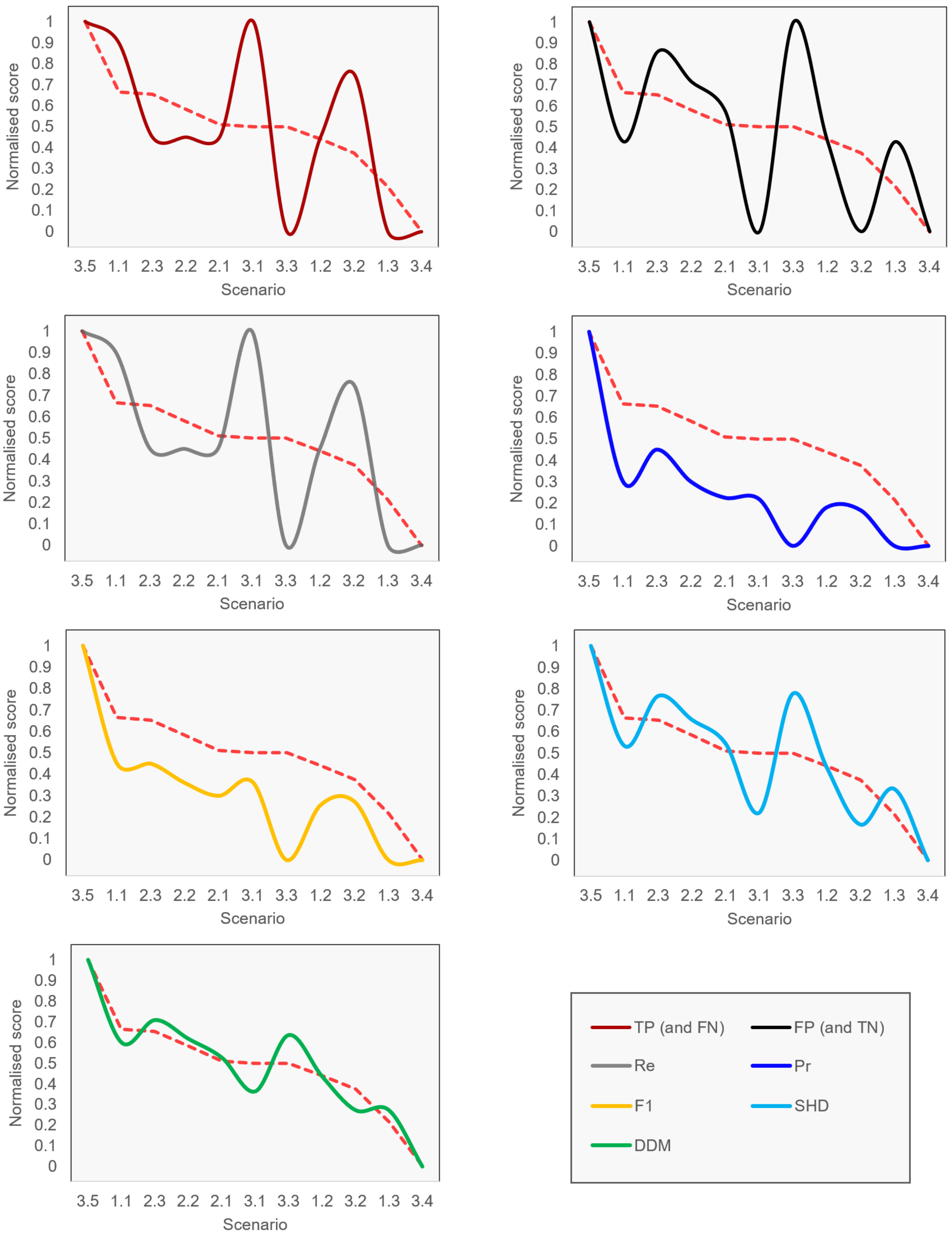

**Fig 1.** How the scores produced by the other metrics compare to the scores produced by the BSF (red dashed line), based on the hypothetical scenarios described in Table 5. All the scores are normalised between 0 and 1 and adjusted so that a score of 1 represents the highest accuracy and a score of 0 the lowest accuracy across all metrics. The scenarios are ordered by BSF score.

The score comparisons illustrate interesting consistencies and inconsistencies between metrics. To begin with, the purpose of this analysis is to investigate whether the metrics agree in determining one graph as being more accurate than another graph, irrespective of the difference in score between metrics. On this basis all the metrics are in agreement when it comes to determining the least and most accurate graphs, since all the metrics consider the graph in scenario 3.5 to be the highest scoring graph and the graph in scenario 3.4 to be the lowest scoring graph (although some metrics consider some of the residual graphs to be equally accurate/inaccurate). However, many of the residual scenarios are judged differently depending on the metric. For example, notice how the structural discrepancy-based metrics SHD and DDM, as well as the FP/TN parameters, consider the graph under scenario 3.3 to be more accurate than many of the preceding graphs, something which contradicts all the other metrics (and this also applies to scenarios 2.3 and 1.3, although to a lower extent).

The graphs also highlight the strong inconsistencies between $Pr$ and $Re$. In fact, the score optimisation has revealed that $Re$ provides information that is no richer than the information provided by the TP parameter, since the two relevant graphs are equivalent in terms of the relative difference in score across the different scenarios. This occurs because by knowing TP, we also know FN. Since $Re$ is solely determined by those two parameters, there is no gain in information over the TP parameter. This observation extends to the F1 score whose relative score difference across the different scenarios appears to be largely determined by $Pr$. While there is a strong association between the normalised scores produced by the different metrics, the key observation here is that each oscillation observed in the scores produced of a metric corresponds to a disagreement with the BSF metric in terms of judging whether the preceding graph is more accurate than the proceeding graph. Overall, none of the other metrics is in full agreement with the ordering of the scenarios as determined by the BSF.

### 5.2. *Empirical analysis*

The empirical analysis is based on the experiments carried in (Constantinou et al., 2020) to validate different BN structure learning algorithms, with and without noisy data. Specifically, 373 graphs are reused in this study to investigate how they are judged by the different metrics. These graphs were generated by 14[3] different algorithms tested in (Constantinou et al., 2020) and are based on synthetic data of varying sample size sampled from six different real-world BN models.

Fig 2 repeats the analysis of Fig 1, although this time excluding TP and FP. In contrast to the hypothetical experiments in subsection 5.1 which assumed a single ground truth graph, the empirical analysis is based on six diverse real-world graphs of varying size and complexity, ranging from 8 to 109 variables and from 8 to 195 true arcs. This, however, creates an issue with the structural discrepancy-based scores of SHD and DDM which are not comparable across graphs of varying sizes. For example, an SHD error of 10 may be good for a graph that consists of 100 variables, but poor for a graph that consists of just 10 variables. This issue is overcome by normalising the SHD and DDM scores with respect to all the scores that correspond to the same true graph, rather than over all scores.

[3] This number excludes the results produced by the RFCI-BSC algorithm tested in (Constantinou et al., 2020). This is because RFCI-BSC is a non-deterministic algorithm that produces a different result with each execution.

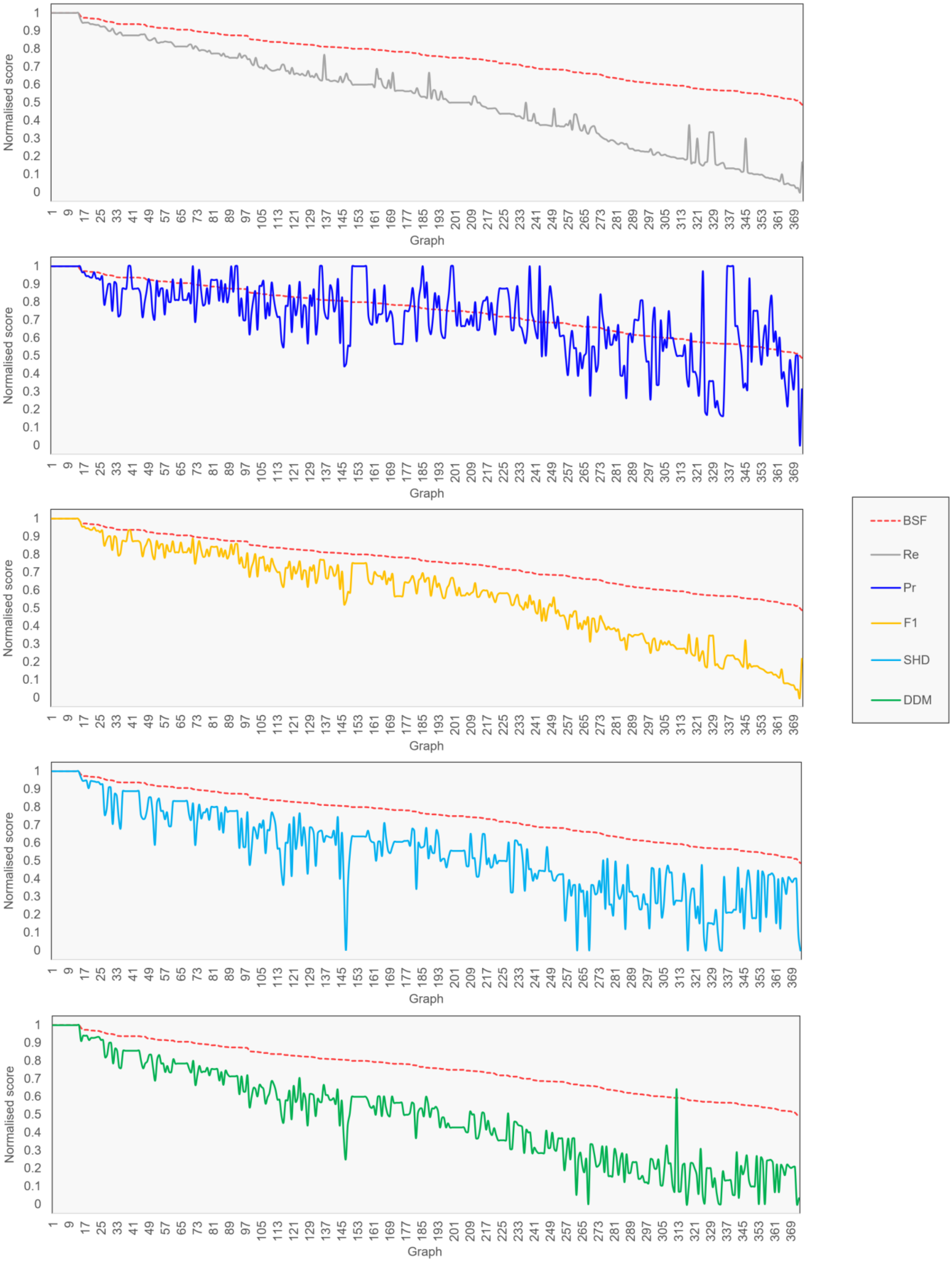

**Fig 2.** How the scores produced by the other metrics compare to the scores produced by the BSF (red dashed line), based on the 373 learnt graphs in (Constantinou et al., 2020). All the scores are normalised between 0 and 1 and adjusted so that a score of 1 represents the highest accuracy and a score of 0 the lowest accuracy across all metrics. The scenarios are ordered by BSF score.

As in subsection 5.1, each subsequent graph on the $x$-axis represents a less accurate graph as determined by the BSF. Once more, each oscillation in the scores corresponds to a disagreement between the metric that produces the oscillation and the BSF, in terms of judging whether one graph is more accurate than another graph. The more aggressive the oscillation the stronger the disagreement. For example, the oscillation produced by SHD (and to a less extent by DDM) at graph 147 may be in agreement with BSF in terms of the graph 147 being less accurate than all the preceding graphs, but contradicts the BSF by also claiming that graph 147 is less accurate than almost all the proceeding graphs. An example with the reverse disagreement of the same scale involves the $Pr$ scores assigned to graphs 337 to 340 which, in contrast to BSF, they are considered to be more accurate than almost all the preceding as well as all the proceeding graphs. Lastly, notice that only the 373$^{rd}$ graph is assigned a BSF score less than 0.5 (less than 0 before normalisation), implying that this is the only graph judged as being less accurate than the baseline empty or fully connected graph. Overall, the results are consistent with the hypothetical analysis and highlight numerous disagreements between metrics.

## 6. Concluding remarks

In the absence of an agreed and an adequate evaluation method it is difficult to reach a consensus about which of the competing algorithms is 'best', or whether one imperfect graph is more accurate than another imperfect graph. The different evaluation methods increase the risk of inconsistency whereby one evaluator determines graph $A$ to be superior to graph $B$, and another evaluator concludes the opposite. In such situations, the selection of an evaluation method is as important as the algorithm itself since the evaluation method practically judges the performance of that algorithm.

Korb and Nicholson (2011) acknowledged that there is no agreed evaluation process due to inadequate experimental survey of the main competing algorithms of the field, and stated that "*every publication in the field attempts to make some kind of empirical case for the particular algorithm being described in that publication*". Motivated by this claim, this paper provided a comprehensive review of the evaluation approaches used in this field of research. The results revealed high inconsistency in the evaluation process across papers, as well as high inconsistency amongst conclusions derived by the different metrics used to evaluate the relationship between a learnt graph and the true graph. These inconsistencies will often lead to conflicting conclusions about which algorithm, or a learnt graph, is 'best'. Importantly, all the well-established metrics that are fully orientated towards graphical discovery were found to be imbalanced in different ways, thereby leading to the observed inconsistencies.

On this basis, this paper proposed a new metric, namely the BSF, that takes into consideration all the confusion matrix parameters and eliminates the score imbalance by adjusting the reward function relative to the difficulty of discovering an edge, or no edge, proportional to their occurrence rate in the ground truth graph. Therefore, the BSF can be used in conjunction with other metrics to offer an alternative and a balanced evaluation about validity of a structure learning algorithm or about the relationship of a learnt graph with respect to a gold standard graph. Specifically, the BSF score ranges from -1 to 1, where -1 corresponds to the most inaccurate graph (the reverse of the true graph), 0 corresponds to an ignorant baseline graph (a fully connected or an empty graph), and 1 to the most accurate graph (matches the true graph). Therefore, the score is meaningful in terms of revealing the likely relative superiority of a learnt graph with respect to a baseline ignorant graph. This is an important feature that not only addresses the inherent biases observed in the other metrics, but also makes the scores produced by the BSF comparable across studies that may be based on graphs of varying complexity.

## Acknowledgements

This research was supported by the ERSRC Fellowship project EP/S001646/1 on Bayesian Artificial Intelligence for Decision Making under Uncertainty, and by The Alan Turing Institute in the UK under the EPSRC grant EP/N510129/1.

## References

Acid, S., and de Campos, L. M. (2003). Searching for Bayesian network structures in the space of restricted acyclic partially directed graphs. *Journal of Artificial Intelligence Research*, vol. 18, pp. 445–490.

Alameddine, I., Cha, Y., and Reckhow, K. H. (2011). An evaluation of automated structure learning with Bayesian networks: An application to estuarine chlorophyll dynamics. *Environmental Modelling & Software*, vol. 26, pp. 163–172.

Aliferis, C. F., Statnikov, A., Tsamardinos, I., Mani, S., and Koutsoukos, X. (2010). Local Causal and Markov Blanket Induction for Causal Discovery and Feature Selection for Classification Part I: Algorithms and Empirical Evaluation. *Journal of Machine Learning Research*, vol. 11., pp. 171–234.

Alonso-Barba, J. I., and delaOssa, L. (2011). Structural learning of Bayesian networks using local algorithms based on the space of orderings. *Statistical Computing*, vol. 15, pp. 1881–1895.

Alonso-Barba, J. I., delaOssa, L., Gamez, J. A., and Puerta, J. M. (2013). Scaling up the Greedy Equivalence Search algorithm by constraining the search space of equivalence classes. *International Journal of Approximate Reasoning*, vol. 54, pp. 429–451.

Andersson, S. A., Madigan, D. and Perlman, M. D. (1997). A characterization of Markov equivalence classes for acyclic digraphs. *Annals of Statistics*, vol. 25, pp. 505–541.

Barlett, M., and Cussens, J. (2013). Advances in Bayesian network learning using integer programming. In *Proceedings of the 29th Conference on Uncertainty in Artificial Intelligence (UAI-2013)*, pp. 182–191, Washington, USA.

Behjati, S., and Beigy, H. (2018). An order-based algorithm for learning structure of Bayesian networks. In *Proceedings of the 9th Conference on Probabilistic Graphical Models (PGM-2018)*, PMLR: vol. 72, pp. 25–36, Prague, Czech Republic.

Borboudakis, G., and Tsamardinos, I. (2013). Scoring and searching over Bayesian networks with causal and associative priors. In *Proceedings of the 29th Conference on Uncertainty in Artificial Intelligence (UAI-2013)*, pp. 102–111, Washington, USA.

Bouchaala, L., Masmoudi, A., Gargouri, F., and Rebai, A. (2010). Improving algorithms for structure learning in Bayesian networks using a new implicit score. *Expert Systems with Applications*, vol. 37, pp. 5470–5475.

Bouckaert, R. R. (1994). Properties of Bayesian Belief Network Learning Algorithms. In *Proceedings of the 10th Conference on Uncertainty in Artificial Intelligence (UAI-1994)*, pp. 102–109, Seattle, Washington, USA.

Brenner, E., and Sontag, D. (2013). SparsityBoost: a new scoring function for learning Bayesian network structure. In *Proceedings of the 29th Conference on Uncertainty in Artificial Intelligence (UAI-2013)*, pp. 112–121, Washington, USA.

Bromberg, F., and Margaritis, D. (2009). Improving the Reliability of Causal Discovery from Small Data Sets Using Argumentation. *Journal of Machine Learning Research*, vol. 10, pp. 141–180.

Buhlmann, P., Kalisch, M., and Maathuis, M. H. (2010). Variable selection in high-dimensional linear models: partially faithful distributions and the PC-simple algorithm. *Biometrika*, vol. 97, pp. 261–278.

Cano, A., Gomez-Olmedo, M., Masegosa, A. R., and Moral, S. (2013). Locally averaged Bayesian Dirichlet metrics for learning the structure and the parameters of Bayesian networks. *International Journal of Approximate Reasoning*, vol. 54, pp. 526–540.

Castelo, R., and Siebes, A. (2000). Priors on network structures. Biasing the search for Bayesian networks. *International Journal of Approximate Reasoning*, vol. 24, pp. 39–57.

Castelo, R., and Kocka, T. (2003). On Inclusion-Driven Learning of Bayesian Networks. *Journal of Machine Learning Research*, vol. 4., pp. 527–574.

Chen, E. Y., Darwiche, A., and Choi, A. (2018) On pruning with the MDL Score. *International Journal of Approximate Reasoning*, vol. 92, pp. 363–375.

Chen, X., Anantha, G., and Xiaotong, L. (2008). Improving Bayesian network structure learning with Mutual Information-Based node ordering in the K2 algorithm. *IEEE Transactions on Knowledge and Data Engineering*, vol. 20, Iss. 5.

Cheng, J., Greiner, R., Kelly, J., Bell, D., and Liu, W. (2002). Learning Bayesian networks from data: An information-theory based approach. *Artificial Intelligence*, vol. 137, pp. 43–90.

Chickering, D. M. (1995). A Transformational Characterization of Equivalent Bayesian Network Structures. In *Proceedings of the 11th Conference on Uncertainty in Artificial Intelligence (UAI-1995)*, pp. 87–98, Quebec, Canada.

Chickering, D. M. (2002). Optimal structure identification with greedy search. *Journal of Machine Learning Research*, pp. 507–554.

Chow, C. K., and Liu, C. N. (1968). Approximating Discrete Probability Distributions with Dependence Trees. *IEEE Transactions on Information Theory*, vol. 14, Iss. 3, pp. 462–467.

Correia, A., Cussens, J., and de Campos, C. P. (2020). On Pruning for Score-Based Bayesian Network Structure Learning. In *Proceedings of the 23rd International Conference on Artificial Intelligence and Statistics (AISTATS)*, PMLR: vol. 108, Palermo, Italy.

Coid, J. W., Ullrich S., Kallis, C., Freestone, M., Gonzalez, R., et al. (2016). Improving risk management for violence in mental health services: a multimethods approach. *Programme Grants for Applied Research*, vol. 4, Iss. 16.

Colombo, D. and Maathuis, M. H. (2014). Order-Independent Constraint-Based Causal Structure Learning. *Journal of Machine Learning Research*, vol. 15, pp. 3921–3962.

Colombo, D., Maathuis, M., Kalisch, M. and Richardson, T. S. (2012). Learning High-Dimensional Directed Acyclic Graphs with Latent and Selection Variables. *The Annals of Statistics*, vol. 40, Iss. 1, pp. 294–321.

Constantinou, A. C. and Fenton, N. (2017). The future of the London Buy-To-Let property market: Simulation with Temporal Bayesian Networks. *PLoS ONE*, vol. 12, Iss. 6, e0179297.

Constantinou, A. C. and Fenton, N. (2018). Things to know about Bayesian Networks. *Significance*, vol. 15, Iss. 2, pp. 19–23.

Constantinou, A. C., Fenton, N. and Neil, M. (2019). How do some Bayesian Network machine learned graphs compare to causal knowledge? *Under review*.

Constantinou, A. C., Liu, Y., Chobtham, K., Guo, Z., and Kitson, N. K. (2020). Large-scale empirical validation of Bayesian Network structure learning algorithms with noisy data. arXiv:2005.09020 [cs.LG]

Cooper, G. F. and Herskovits, E. (1992). A Bayesian method for the induction of probabilistic networks from data. *Machine Learning*, vol. 9, pp. 309–347.

Córdoba I., Garrido-Merchán E.C., Hernández-Lobato D., Bielza C., and Larrañaga P. (2018) Bayesian Optimization of the PC Algorithm for Learning Gaussian Bayesian Networks. In: Herrera F. et al. (eds) *Advances in Artificial Intelligence. CAEPIA 2018. Lecture Notes in Computer Science*, vol 11160. Springer, Cham.

Cussens, J. (2008). Bayesian network learning by compiling to weighted MAX-SAT. In: McAllester, David and Myllymaki, Petri, (eds.) *Proceedings of the 24th Conference on Uncertainty in Artificial Intelligence (UAI-2008)*, Corvallis, Oregon, pp. 105–112.

Cussens, J. (2011). Bayesian network learning with cutting planes. In *Proceedings of the 27th Conference on Uncertainty in Artificial Intelligence* (UAI-2011), pp. 153–160, Barcelona, Spain.

de Campos, C. P., and Ji, Q. (2011). Efficient Structure Learning of Bayesian Networks using Constraints. *Journal of Machine Learning Research*, vol. 12, pp. 663–689.

de Campos, C. P., Scanagatta, M., Corani, G., and Zaffalon, M. (2018). Entropy-based pruning for learning Bayesian networks using BIC. *Artificial Intelligence*, vol. 260, pp. 42–50.

de Jongh, M. and Druzdzel, M. J. (2009). A comparison of structural distance measures for causal Bayesian network models. In *M. Klopotek, A. Przepiorkowski, S. T. Wierzchon, and K. Trojanowski, editors, Recent Advances in Intelligent Information Systems, Challenging Problems of Science, Computer Science series*, Academic Publishing House EXIT, pp. 443–456.

de Morais, S. R., and Aussem, A. (2010). An efficient and scalable algorithm for local Bayesian network structure discovery. In *Balcázar J.L., Bonchi F., Gionis A., Sebag M. (eds) Machine Learning and Knowledge Discovery in Databases. ECML PKDD 2010. Lecture Notes in Computer Science*, vol 6323. Springer, Berlin, Heidelberg.

Daly, R., and Shen, Q. (2009). Learning Bayesian network equivalence classes with ant colony optimization. *Journal of Artificial Intelligence Research*, vol. 35, pp. 391–447.

Danks D., Glymour C., and Tillman R. E. (2009). Integrating locally learned causal structures with overlapping variables. In *Proceedings of the 22nd Conference on Advances in Neural Information Processing Systems 21 (NIPS 2008)*, pp. 1665–72, Vancouver, BC, Canada.

Dash, D., and Druzdzel, M. J. (2003). Robust Independence Testing for Constraint-Based Learning of Causal Structure. In *Proceedings of the 19th Annual Conference on Uncertainty in Artificial Intelligence (UAI-2003)*, pp. 167–174, San Francisco, CA

Dash, D., and Cooper, G. (2004). Model averaging for prediction with Discrete Bayesian networks. *Journal of Machine Learning Research*, vol. 5, pp. 1177–1203.

Djordjilovic, V., Chiogna, M., and Vomlel, J. (2017). An empirical comparison of popular structure learning algorithms with a view to gene network inference. *International Journal of Approximate Reasoning*, vol. 88, pp. 602–613.

Eaton, D., and Murphy, K. (2007). Bayesian structure learning using dynamic programming and MCMC. In *Proceedings of the 23rd Annual Conference on Uncertainty in Artificial Intelligence (UAI-07)*, pp. 101–108, Vancouver, BC, Canada.

Eggeling, R., Viinikka, J., Vuoksenmaa, A., and Koivisto, M. (2019). On structure priors for learning Bayesian networks. In *Proceedings of the 22nd International Conference on Artificial Intelligence and Statistics (AISTATS)*, PCML: vol. 89. Naha, Okinawa, Japan.

Elidan, G., and Gould, S. (2008). Learning Bounded Treewidth Bayesian Networks. *Journal of Machine Learning Research*, vol. 9, pp. 2699–2731.

Ellis, B., and Wong, W. H. (2008). Learning Causal Bayesian Network Structures from Experimental Data. *Journal of the American Statistical Association*, vol. 103, Iss. 482, pp. 778–789

Etxeberria, R., Larranaga, P., and Picaza, J. M. (1997). Analysis of the behaviour of generic algorithms when learning Bayesian network structure from data. *Pattern Recognition Letters*, vol. 18, pp. 1269–1273.

Fenton, N., Neil, M., Lagnado, D., Marsh, W., Yet, B., et al. (2016). How to model mutually exclusive events based on independent causal pathways in Bayesian network models. *Knowledge-Based Systems*, vol. 113, pp. 39–50.

Fenton, N., and Neil, M. (2018). Risk assessment and Decision Analysis with Bayesian Networks (Second Edition). Chapman and Halll/CRC Press.

Flores, M. J., Nicholson, A. E., Brunskill, A., Korb, K. K., and Mascaro, S. (2011). Incorporating expert knowledge when learning Bayesian network structure: A medical case study. *Artificial Intelligence in Medicine*, vol. 53, pp. 181–204.

Friedman, N., Nachman, I. and Peer, D. (1999). Learning Bayesian network structure from massive datasets: the 'sparse candidate' algorithm. In *Proceedings of the 15th Conference on Uncertainty in Artificial Intelligence (UAI-1999)*, pp. 206–215, Stockholm, Sweden.

Friedman, N., and Koller, D. (2003). Being Bayesian about network structure. A Bayesian approach to structure discovery in Bayesian networks. *Machine Learning*, vol. 50, pp. 95–125.

Gao, T., Fadnis, K., and Campbell, M. (2017). Local-to-Global Bayesian network structure learning. In *Proceedings of the 34th International Conference on Machine Learning*, PMLR: vol. 70, Sydney, Australia.

Gao, T., and Wei, D. (2018). Parallel Bayesian Network Structure Learning. In *Proceedings of the 35th International Conference on Machine Learning*, PMLR: vol. 80, Stockholm, Sweden.

Gasse, M., Aussem, A., and Elghazel, H. (2012). An Experimental Comparison of Hybrid Algorithms for Bayesian Network Structure Learning. *Lecture Notes in Computer Science, Springer, 2012, Machine Learning and Knowledge Discovery in Databases*, 7523, pp.58–73

Gheisari, S., and Meybodi, M. R. (2016). BNC-PSO: structure learning of Bayesian networks by Particle Swarm Optimization. *Information Sciences*, vol. 348, pp. 272–289.

Goudie, R. J. B., and Mukherjee, S. (2016). A Gibbs Sampler for Learning DAGs. *Journal of Machine Learning Research*, vol. 17, pp. 1–39.

Grzegorczyk, M, and Husmeier, D. (2008). Improving the structure MCMC sampler for Bayesian networks by introducing a new edge reversal move. *Machine Learning*, vol. 71, pp. 265–305.

Hand, D., and Christen, P. (2018). A note on using the F-measure for evaluating record linkage algorithms. *Statistics and Computing*, vol. 28, Iss. 3, pp. 539–547.

Harris, N., and Drton, M. (2013). PC Algorithm for Nonparanormal Graphical Models. *Journal of Machine Learning Research*, vol. 14, pp. 3365–3383.

He, R., Tian, J., and Wu, H. (2016). Structure Learning in Bayesian Networks of a Moderate Size by Efficient Sampling. *Journal of Machine Learning Research*, vol. 17, pp. 1–54.

Heckerman, D., Geiger, D., and Chickering, D. M. (1995). Learning Bayesian Networks: The Combination of Knowledge and Statistical Data. *Machine Learning*, vol. 20, pp. 197–243.

Huang, S., Li, J., Ye, J., Fleisher, A., Chen, K., Wu, T., and Reiman, E. (2013). A Sparse Structure Learning Algorithm for Gaussian Bayesian Network Identification from High-Dimensional Data. *IEEE Transactions on Pattern Analysis and Machine Intelligence*, vol. 25, Iss. 6, pp. 1328–1342.

Jaakkola, T., Sontag, D., Globerson, A., and Meila, M. (2010). Learning Bayesian Network Structure using LP Relaxations. In *Proceedings of the 13th International Conference on Artificial Intelligence and Statistics (AISTATS)*, PMLR: vol. 9, Sardinia, Italy.

Jennings, D., and Corcoran, J. N. (2018). A birth and death process for Bayesian network structure inference. *Probability in the Engineering and Informational Sciences*, vol. 32, Iss. 4, pp. 615–625.

Ji, J., Wei, H., and Liu, C. (2013). An artificial bee colony algorithm for learning Bayesian networks. *Statistics and Computing*, vol. 17, pp. 983–994.

Ji, J., Yang, C., Liu, J., Liu, J., and Yin, B. (2017). A comparative study on swarm intelligence for structure learning of Bayesian networks. *Soft Computing*, vol. 21, pp. 6713–6738.

Kalisch, M., and Buhlmann, P. (2007). Estimating High-Dimensional Directed Acyclic Graphs with the PC-Algorithm. *Journal of Machine Learning Research*, vol. 8, pp. 613–636.

Kiattikun, C., and Constantinou, A. C. (2020). Bayesian network structure learning with causal effects in the presence of latent variables. In *Proceedings of the 10th International Conference on Probabilistic Graphical Models (PGM2020)*, PMLR, Aalborg, Denmark.

Kitson, N. K., and Constantinou, A. C. (2020). Learning Bayesian networks from demographic and health survey data. *arXiv:1912.00715* [cs.AI].

Koivisto, M., and Sood, K. (2004). Exact Bayesian structure discovery in Bayesian networks. *Journal of Machine Learning Research*, vol. 5, pp. 549–573.

Korb, K. and Nicholson, A. (2011). Bayesian Artificial Intelligence (Second Edition). London, UK: CRC Press.

Lee, C., and van Beek, P. (2017). An experimental analysis of anytime algorithms for Bayesian network structure learning. In *Proceedings of the 3rd International Workshop on Advanced Methodologies for Bayesian Networks*, PMLR: vol. 73, pp. 69–80, Kyoto, Japan.

Li, A. C., and van Beek, P. (2018). Bayesian network structure learning with side constraints. In *Proceedings of the 9th Conference on Probabilistic Graphical Models (PGM-2018)*, PMLR: vol. 72, pp. 225–236, Prague, Czech Republic.

Lagani, V., Triantafillou, S., Ball, G., Tegnér, J., and Tsamardinos, I. (2016). Probabilistic Computational Causal Discovery for Systems Biology. In *Geris L., Gomez-Cabrero D. (eds) Uncertainty in Biology. Studies in Mechanobiology, Tissue Engineering and Biomaterials*, vol. 17, Springer, Cham.

Larranaga, P., Poza, M., Yurramendi, Y., Murga, R. H., and Kuijpers, C. M. H. (1996). Structure learning of Bayesian networks by Genetic Algorithms: A performance analysis of control parameters. *IEEE Transactions on Pattern Analysis and Machine Intelligence*, vol. 18, Iss. 9, pp. 912-926.

Liao, Z., Sharma, C., Cussens, J., and Peter v. B. (2019). Finding all Bayesian network structures within a factor of optimal. In *Proceedings of the 33rd AAAI Conference on Artificial Intelligence (AAAI-19)*, Hawaii, USA.

Maathuis, M. H., Colombo, D. K. M. and Bühlmann, P. (2010). Predicting causal effects in large-scale systems from observational data. *Nature Methods*, vol. 7, pp. 247–248.

Maathuis, M. H., Kalisch, M., and Buhlmann, P. (2009). Estimating High-Dimensional Intervention Effects from Observational Data. *The Annals of Statistics*, vol. 37, Iss. 6A, pp. 3133–3164.

Madigan, D., Andersson, S. A., Perlman, M. D., and Volinsky, C. T. (1996). Bayesian model averaging and model selection for Markov Equivalence Classes of Acyclic Digraphs. *Communications in Statistics – Theory and Methods*, vol. 25, Iss. 11.

Malone, B., Jarvisalo, M., and Myllymaki, P. (2015). Impact of learning strategies on the quality of Bayesian networks: An empirical evaluation. In *Proceedings of the 31st Conference of Uncertainty in Artificial Intelligence (UAI-2015)*, Corvallis, Oregon.

Margaritis, D. (2003). Learning Bayesian Network Model Structure from Data. Pittsburgh, PA: Carnegie Mellon University.

Masegosa, A., and Moral, S. (2013). An interactive approach for Bayesian network learning using domain/expert knowledge. *International Journal of Approximate Reasoning*, vol. 54, pp. 1168–1181.

Meek, C. (1995). Causal inference and causal explanation with background knowledge. In *Proceedings of the 11th Conference on Uncertainty in Artificial Intelligence (UAI-1995)*, pp. 403–410.

Meganck S., Leray P., Manderick B. (2006). Learning Causal Bayesian Networks from Observations and Experiments: A Decision Theoretic Approach. In*: Torra V., Narukawa Y., Valls A., Domingo-Ferrer J. (eds) Modeling Decisions for Artificial Intelligence. MDAI 2006. Lecture Notes in Computer Science*, vol. 3885, Berlin, Heidelberg

Moore, A. and Wong, W. K. (2003). Optimal Reinsertion: A new search operator for accelerated and more accurate Bayesian network structure learning. In *Proceedings of the 20th International Conference on Machine Learning (ICML-2003)*, Washington DC.

Myers, J. W., Laskey, K. B., and Levitt, T. (1999). Learning Bayesian networks from incomplete data with stochastic search algorithms. In *Proceedings of the 15th Conference on Uncertainty in Artificial Intelligence (UAI-1999)*, pp. 476–485, Stockholm, Sweden.

Nandy, P., Hauser, A., and Maathuis, M. (2018). High-dimensional consistency in score-based and hybrid structure learning. The Annals of Statistics, vol. 46, Iss. 6A, pp. 3151–3183.

Needham, C. J., Bradford, J. R., Bulpitt, A. J. and Westhead, D. R. (2007). A Primer on Learning in Bayesian Networks for Computational Biology. *PLoS Computational Biology*, vol. 3, Iss. 8, p. e129.

Nie, S., de Campos, C. P., and Ji, Q. (2017). Efficient learning of Bayesian networks with bounded tree-width. *International Journal of Approximate Reasoning*, vol. 80, pp. 412–427.

Niinimaki, T., and Koivisto, M. (2013). Annealed Importance Sampling for Structure Learning in Bayesian Networks. In *Proceedings of the 23rd International Joint Conference on Artificial Intelligence (IJCAI-2013)*, pp. 1579–1585, Beijing, China.

Noble, J. M. (2010). An algorithm for learning the essential graph. *arXiv:1007.2656 [stat.CO]*

Oates, C. J., Smith, J. Q., Mukherjee, S., and Cussens, J. (2016). Exact estimation of multiple directed acyclic graphs. *Statistics and Computing*, vol. 26, pp. 797–811.

Ogarrio, J. M., Spirtes, P., and Ramsey, J. (2016). A Hybrid Causal Search Algorithm for Latent Variable Models. In *Proceedings of the 8th International Conference on Probabilistic Graphical Models (PGM-2016)*, PMLR: vol. 52, Lugano, Switzerland.

Parviainen, P., Farahani, H. S., and Lagergren, J. (2014). Learning Bounded Tree-width Bayesian Networks using Integer Linear Programming. In *Proceedings of the 17th International Conference on Artificial Intelligence and Statistics (AISTATS-2014)*, Reykjavik, Ireland.

Parviainen, P., and Kaski, S. (2017). Learning structures of Bayesian networks for variable groups. *International Journal of Approximate Reasoning*, vol. 88, pp. 101–127.

Pearl, J. (2013). Causality: Models, Reasoning, and inference (2nd Edition). Los Angeles: Cambridge University Press.

Pellet, J., and Elisseeff, A. (2008). Using Markov Blankets for Causal Structure Learning. (2008). *Journal of Machine Learning Research*, vol. 9, pp. 1295–1342.

Pensar, J., Nyman, H., Lintusaari, J., and Corander, J. (2016). The role of local partial independence in learning of Bayesian networks. *International Journal of Approximate Reasoning*, vol. 69, pp. 91–105.

Pernkopf, F., and Bilmes, J. A. (2010). Efficient heuristics for Discriminative Structure Learning of Bayesian network classifiers. *Journal of Machine Learning Research*, vol. 11, pp. 2323–2360.

Perrier, E., Imoto, S., and Miyano, S. (2008). Finding Optimal Bayesian Network Given a Super-Structure. *Journal of Machine Learning Research*, vol. 9, pp. 2251–2286.

Peters, J., and Buhlmann, P. (2014). Structural Intervention Distance (SID) for Evaluating Causal Graphs. *arXiv:1306.1043v2 [stat.ML]*.

Pinto, P. P., Nagele, A., Dejori, M., Runkler, T. A. and Sousa, J. M. C. (2009). Using a local discovery Ant algorithm for Bayesian network structure learning. *IEEE Transactions on Evolutionary Computation*, vol. 13, Iss. 4, pp. 767–779.

Ramsey, J. (2016). Improving Accuracy and Scalability of the PC Algorithm by Maximizing P-value, Pittsburgh, PA: Center for Causal Discovery.

Ramsey, J. D. (2015). Scaling up Greedy Causal Search for Continuous Variables, Pittsburgh, PA: Center for Causal Discovery.

Ramsey, J., Glymour, M., Sanchez-Romero, R. and Glymour, C. (2017). A million variables and more: the Fast Greedy Equivalence Search algorithm for learning high-dimensional graphical causal models, with an application to functional magnetic resonance images. *International Journal of Data Science and Analytics*, vol. 3, Iss. 2, pp. 121–129.

Ramsey, J., Zhang, J., and Spirtes, P. (2006). Adjacency-Faithfulness and Conservative Causal Inference. In *Proceedings of the 22nd Conference on Uncertainty in Artificial Intelligence*, Oregon, USA.

Richardson, T. (1996). A Discovery Algorithm for Directed Cyclic Graphs. In *Proceedings of the 12th International Conference on Uncertainty in Artificial Intelligence (UAI-1996)*, pp. 454–461, Oregon, USA.

Sachs, K., Perez, O., Peer, D., Lauffenburger, D. A., and Nolan, G. P. (2005). Causal Protein-Signaling Networks Derived from Multiparameter Single-Cell Data. *Science*, vol. 308, Iss. 5721, pp. 523–529.

Sanchez-Romero, R., Ramsey, J. D., Zhang, K., Glymour, M. R. K., Huang, B., and Glymour, C. (2018). Causal Discovery of Feedback Networks with Functional Magnetic Resonance Imaging. *bioRxiv*: https://doi.org/10.1101/245936.

Scanagatta, M., Corani, G., de Campos, C. P, and Zaffalon, M. (2015). Learning Treewidth-Bounded Bayesian Networks with Thousands of Variables. In *Advances in Neural Information Processing Systems*, pp. 1462–1470.

Scanagatta, M., Corani, G., de Campos, C. P. and Zaffalon, M. (2018). Approximate structure learning for large Bayesian networks. *Machine Learning*, vol. 107, Iss. 8-10, pp. 1209–1227.

Schmidt, M., Niculescu-Mizil, A., and Murphy, K. (2007). Learning graphical model structure using L1-regularization paths. In *Proceedings of the 22nd National Conference on Artificial Intelligence (AAAI-2007)*, vol. 2, pp. 1278–1283, Vancouver, BC, Canada.

Scutari, M. and Brogini, A. (2012). Bayesian Network Structure Learning with Permutation Tests. *Communications in Statistics - Theory and Methods*, vol. 41, Iss. 16-17, pp. 3233–3243.

Scutari, M. (2016). An Empirical-Bayes Score for Discrete Bayesian Networks. In *Proceedings of 8th Conference on Probabilistic Graphical Models (PGM-2016)*, PMLR: vol. 52, pp. 438–448, Lugano, Switzerland.

Scutari, M., Graafland, C. E., and Guitierrez, J. M. (2018). Who Learns Better Bayesian Network Structures: Constraint-Based, Score-based or Hybrid Algorithms? In *Proceedings of the 9th Conference on Probabilistic Graphical Models (PGM-2018)*, PMLR: vol. 72, pp. 416–427, Prague, Czech Republic.

Shimizu, S., Hyvarinen, A., Hoyer, P. O., and Kano, Y. (2006). Finding a causal ordering via independent component analysis. *Computational Statistics & Data Analysis*, vol. 50, pp. 3278–3293.

Sierra, B., Lazkano, E., Jauregi, E., and Irigoien, I. (2009). Histogram distance-based Bayesian network structure learning: A supervised classification specific approach. *Decision Support Systems*, vol. 48, pp. 180–190.

Silander, T., Leppa-aho, J., Jaasaari, E., and Roos, T. (2018). Quotient Normalized Maximum Likelihood Criterion for Learning Bayesian Network Structures. In *Proceedings of the 21st International Conference on Artificial Intelligence and Statistics (AISTATS-2018)*, PMLR: vol. 84, pp. 948–957, Canary Islands, Spain.

Singh, M., and Valtorta, M. (1995). Construction of Bayesian network structures from data: A brief survey and an efficient algorithm. *International Journal of Approximate Reasoning*, vol. 12, pp. 111–131.

Smith, S. M., Miller, K. L., Salimi-Khorshidi, G., Webster, M., Beckmann, C. F., Nichols, T. E., Ramsey, J. D., and Woolrich, M. W. (2011). Network modelling methods for FMRI. *NeuroImage*, vol. 54, pp. 875–891.

Smit, N. M., Lagnado, D. A., Morgan, R. M., and Fenton, N. E. (2016). Using Bayesian networks to guide the assessment of new evidence in an appeal case. *Crime Science*, vol. 5, Iss. 9, pp. 1–12.

Spirtes, P. and Glymour, C. (1991). An Algorithm for Fast Recovery of Sparse Causal Graphs. *Social Science Computer Review*, vol. 9, Iss. 1.

Spirtes, P., Meek, C., and Richardson, T. (1995). Causal Inference in the Presence of Latent Variables and Selection Bias. In *Proceedings of the 11th Conference on Uncertainty in Artificial Intelligence (UAI-1995)*, pp. 499–506, Quebec, Canada.

Spirtes, P., Glymour, C. and Scheines, R. (2000). Causation, Prediction, and Search: 2nd Edition. Cambridge, Massachusetts, London, England: The MIT Press.

Spirtes, P. and Meek, C. (1995). Learning Bayesian networks with discrete variables from data. In *Proceedings of the 1st Annual Conference on Knowledge Discovery and Data Mining*, Montreal, Quebec, Canada.

Srinivas, S., Russell, S., and Agogino, A. (1990). Automated construction of sparse Bayesian networks from unstructured probabilistic models and domain information. *Machine Intelligence and Pattern Recognition*, vol. 10, pp. 295–308.

Studeny, M., Vomlel, J., and Hemmecke, R. (2010). A geometric view on learning Bayesian network structures. *International Journal of Approximate Reasoning*, vol. 51, pp. 573–586.

Studeny, M. and Haws, D. (2014). Learning Bayesian network structure: Towards the essential graph by integer linear programming tools. *International Journal of Approximate Reasoning*, vol. 55, pp. 1043–1071.

Suzuki, J. (2015). Consistency of Learning Bayesian Network Structures with Continuous Variables: An Information Theoretic Approach. *Entropy*, vol. 17, pp. 5752–5770.

Suzuki, J. (2017). A novel Chow-Liu algorithm and its application to gene differential analysis. *International Journal of Approximate Reasoning*, vol. 80, pp. 1–18.

Tabar, V. R. (2017). A simple node ordering method for the K2 algorithm based on the factor analysis. In *Proceedings of the 6th International Conference on Pattern Recognition Applications and Methods*, vol. 1: ICPRAM, pp. 273–280, Porto, Portugal.

Talvitie, T., Eggeling, R., and Koivisto, M. (2018). Finding optimal Bayesian networks with local structure. In *Proceedings of the 9th Conference on Probabilistic Graphical Models (PGM-2018)*, PMLR: vol. 72, pp. 451–462, Prague, Czech Republic.

Teyssier, M., and Koller, D. (2005). Ordering-based search: a simple and effective algorithm for learning Bayesian networks. In *Proceedings of the 21st Conference on Uncertainty in Artificial Intelligence (UAI-2005)*, pp. 584–590, Scotland, UK.

Tian, J. (2000). A Branch-and-Bound algorithm for MDL learning Bayesian networks. In *Proceedings of the 16th Conference on Uncertainty in Artificial Intelligence (UAI-2000)*, pp. 580–588, California, USA.

Triantafillou, S. and Tsamardinos, I. (2016). Score based vs constraint based causal learning in the presence of confounders. In *Proceedings of the 32nd Conference on Uncertainty in Artificial Intelligence*, New York, USA.

Tsamardinos, I., Aliferis, C. and Statnikov, A. (2003). Algorithms for Large Scale Markov Blanket. In *Proceedings of the 16th Florida Artificial Intelligence Research Society Conference*, Florida, USA.

Tsamardinos, I., Brown, L. E. and Aliferis, C. F. (2006). The Max-Min Hill-Climbing Bayesian Network Structure Learning Algorithm. *Machine Learning*, vol. 65, pp. 31–78.

Tsirlis, K., Lagani, V., Triantafillou, S. and Tsamardinos, I. (2018). On scoring Maximal Ancestral Graphs with the Max–Min Hill Climbing algorithm. *International Journal of Approximate Reasoning*, vol. 102, pp. 74–85.

Ueno, M. (2011). Robust learning Bayesian networks for prior belief. In *Proceedings of the 27th Conference on Uncertainty in Artificial Intelligence (UAI-2011)*, pp. 698–707, Barcelona, Spain.

Verma, T. S., and Pearl, J. (1990). Equivalence and synthesis of causal models. In *Proceedings of the 6th Conference on Uncertainty in Artificial Intelligence (UAI-1990)*, pp. 220–227, Cambridge, MA, USA.

Wang, T., and Yang, J. (2010). A heuristic method for learning Bayesian networks using discrete particle swarm optimization. *Knowledge and Information Systems*, vol. 24, pp. 269–281.

Wong, M. L., Lam, W., and Leung, S. (1999). Using Evolutionary Programming and Minimum Description Length Principle for Data Mining of Bayesian Networks. *IEEE Transactions on Pattern Analysis and Machine Intelligence*, vol. 21, Iss. 2, pp. 174–178.

Wong, M. L., and Guo, Y. Y. (2008). Learning Bayesian networks from incomplete databases using a novel evolutionary algorithm. *Decision Support Systems*, vol. 45, pp. 368–383.

Yang, C., Ji, J., Liu, J., Liu, J., and Yin, B. (2016). Structural learning of Bayesian networks by bacterial foraging optimization. *International Journal of Approximate Reasoning*, vol. 69, pp. 147–167.

Yang, J., Li, L., and Wang, A. (2011). A partial correlation-based Bayesian network structure learning algorithm under linear SEM. *Knowledge-Based Systems*, vol. 24, pp. 963–976.

Yaramakala, S. and Margaritis, D. (2005). Speculative Markov Blanket Discovery for Optimal Feature Selection. In *Proceedings of the 5th IEEE International Conference on Data Mining (ICDM)*, Washington DC, USA.

Yehezkel, R., and Lerner, B. (2009). Bayesian network structure learning by recursive autonomy identification. *Journal of Machine Learning Research*, vol. 10, pp. 1527–1570.

Yet, B., Constantinou, A., Fenton, N., Neil, M., Luedeling, E., and Shepherd, K. (2016). A Bayesian Network Framework for Project Cost, Benefit and Risk Analysis with an Agricultural Development Case Study. (2016). A Bayesian Network Framework for Project Cost, Benefit and Risk Analysis with an Agricultural Development Case Study. *Expert Systems with Applications*, vol. 60, pp. 141–155.

Yu, Y., Chen, J., Gao, T., and Yu, M. (2019). DAG-GNN: DAG Structure Learning with Graph Neural Networks. In *Proceedings of the 36th International Conference on Machine Learning, PMLR vol. 97*, Long Beach, California.

Zhang, J. (2008). On the completeness of orientation rules for causal discovery in the presence of latent confounders and selection bias. *Artificial Intelligence*, vol. 172, Iss. 16–17, pp. 1873-1896.

Zhang, Y., Zhang, W., and Xie, Y. (2013). Improved heuristic equivalent search algorithm based on Maximal Information Coefficient for Bayesian Network Structure Learning. *Neurocomputing*, vol. 117, pp. 186–195.

Zhao, J., and Ho, S. (2019). Improving Bayesian network local structure learning via data-driven symmetry correction methods. *International Journal of Approximate Reasoning*, vol. 107, pp. 101–121.

Zuk, O., Margel, S., and Domany, E. (2006). On the number of samples needed to learn the correct structure of a Bayesian network. In *Proceedings of the 22nd Conference on Uncertainty in Artificial Intelligence (UAI-2006)*, pp. 560–567, Cambridge, MA, USA.

Xie, X., and Geng, Z. (2008). A recursive method for structural learning of directed acyclic graphs. *Journal of Machine Learning Research*, vol. 9, pp. 459–483.